%% file: latex/acl_latex.tex
\pdfoutput=1

\documentclass[11pt]{article}

\usepackage[preprint]{acl}

\usepackage{times}
\usepackage{latexsym}

\usepackage[T1]{fontenc}

\usepackage[utf8]{inputenc}

\usepackage{microtype}

\usepackage{inconsolata}

\usepackage{graphicx}
\usepackage{pifont}
\usepackage{multirow}
\usepackage{multicol}
\usepackage{booktabs}
\usepackage{fontawesome}
\usepackage{amsmath}
\usepackage{enumitem}
\usepackage{float}
\usepackage{subcaption}
\usepackage{algorithm}
\usepackage{algorithmic}
\usepackage{xcolor}

\usepackage{stfloats}
\usepackage{amsmath}
\usepackage{amssymb}

\usepackage{graphicx}
\newcommand{\faHuggingFace}{\includegraphics[height=1em]{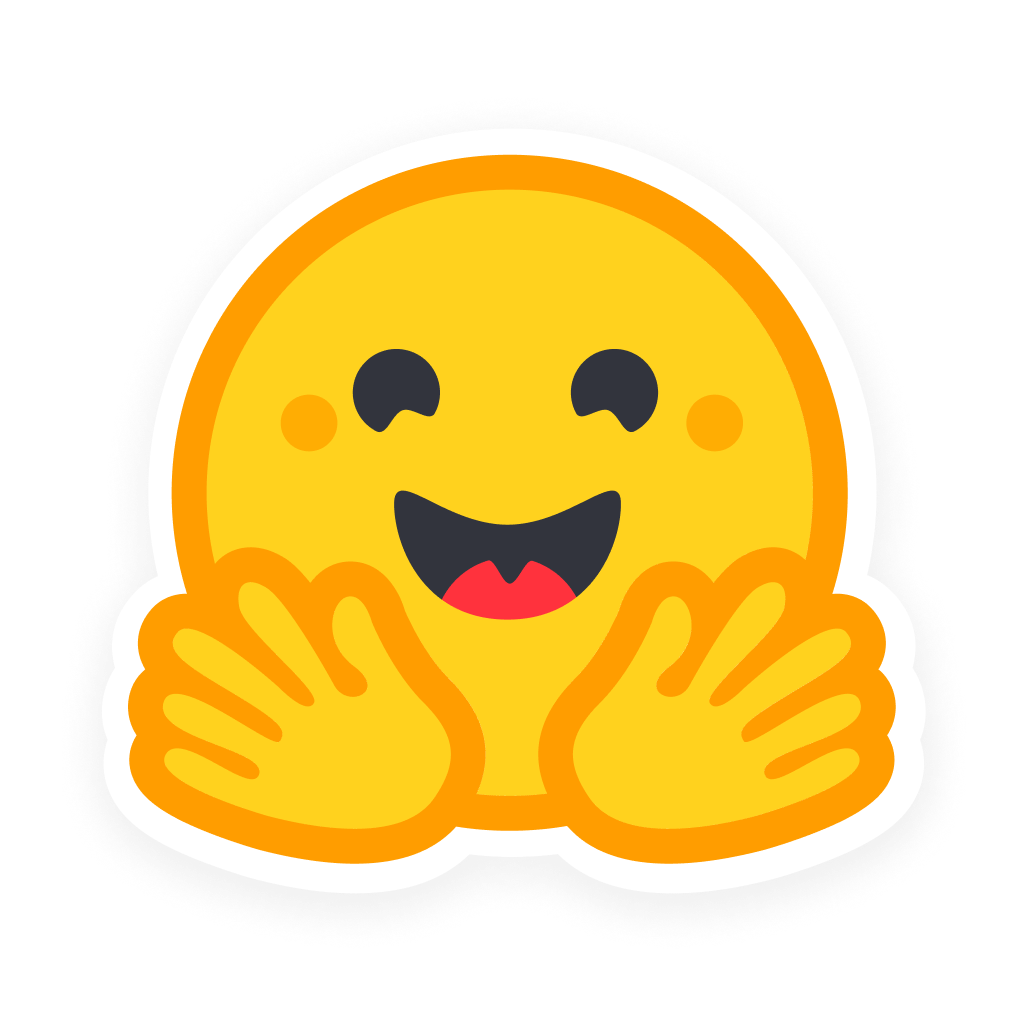}}

\newcommand{\kgw}{{KGW}}
\newcommand{\ie}{{IE}}

\newcommand{\sweet}{{SWEET}}
\newcommand{\logp}{{Log P(X)}}
\newcommand{\logr}{{LogRank}}
\newcommand{\detectgpt}{{DetectGPT}}
\newcommand{\detectb}{{DetectGPT(T5-3B)}}
\newcommand{\gptzero}{{GPTZero}}
\newcommand{\openai}{{OpenAI Classifier}}

\definecolor{third}{HTML}{FFE5D9}
\definecolor{second}{HTML}{FFD7BA} 
\definecolor{best}{HTML}{FEC89A} 

\title{Invisible Entropy: Towards Safe and Efficient\\ Low-Entropy LLM Watermarking}

\author{
 \textbf{Tianle Gu\textsuperscript{1,2}\thanks{{}{} Work done during internship at MBZUAI.}\thanks{{}{} Equal Contribution}},
 \textbf{Zongqi Wang \textsuperscript{2}$\footnotemark[2]$},
 \textbf{Kexin Huang\textsuperscript{3}},
 \textbf{Yuanqi Yao\textsuperscript{4}},
\\
 \textbf{Xiangliang Zhang \textsuperscript{5}\thanks{{}{} Corresponding Authors}},
 \textbf{Yujiu Yang\textsuperscript{2}$\footnotemark[3]$},
 \textbf{Xiuying Chen\textsuperscript{1}$\footnotemark[3]$}
\\
 \textsuperscript{1}Mohamed bin Zayed University of Artificial Intelligence (MBZUAI),\\
 \textsuperscript{2}Tsinghua Shenzhen International Graduate School, Tsinghua University, \\
\textsuperscript{3}Fudan University,\\
 \textsuperscript{4}Shanghai Artificial Intelligence Laboratory,\\
 \textsuperscript{5}University of Notre Dame
\\
}

\begin{document}
\maketitle

\begin{abstract}
\input{sections/01_abstract}
\end{abstract}
\input{sections/02_introduction}
\input{sections/03_related_work}
\input{sections/04_preliminaries}
\input{sections/05_methods}

\input{sections/06_experiments}
\input{sections/07_analysis_and_discussions}
\input{sections/08_conclusion}
\input{sections/09_limitations}

\bibliography{custom}
\clearpage
\appendix
\input{appendicies/sample_for_navigator}
\input{appendicies/algorithms_for_watermark_generation_and_detection}

\input{appendicies/training_details_of_entropy_tagger}
\input{appendicies/implementation_details}
\input{appendicies/detection_time}
\input{appendicies/algoriths_for_threshold_navigator}
\input{appendicies/analysis_of_low_entropy_token}
\end{document}

%% file: sections/01_abstract.tex
Logit-based LLM watermarking traces and verifies AI-generated content by maintaining green and red token lists and increasing the likelihood of green tokens during generation.
However, it fails in low-entropy scenarios, where predictable outputs make green token selection difficult without disrupting natural text flow. 
Existing approaches address this by assuming access to the original LLM to calculate entropy and selectively watermark high-entropy tokens.
However, these methods face two major challenges: (1) high computational costs and detection delays due to reliance on the original LLM, and (2) potential risks of model leakage.
To address these limitations, we propose Invisible Entropy (IE), a watermarking paradigm designed to enhance both safety and efficiency. 
Instead of relying on the original LLM, IE introduces a lightweight feature extractor and an entropy tagger to predict whether the entropy of the next token is high or low. 
Furthermore, based on theoretical analysis, we develop a threshold navigator that adaptively sets entropy thresholds. 
It identifies a threshold where the watermark ratio decreases as the green token count increases, enhancing the naturalness of the watermarked text and improving detection robustness.
Experiments on HumanEval and MBPP datasets demonstrate that IE reduces parameter size by 99\% while achieving performance on par with state-of-the-art methods.
Our work introduces a safe and efficient paradigm for low-entropy watermarking.
\faGithub  \href{https://github.com/Carol-gutianle/IE}{IE-official-repo} \faHuggingFace \href{https://huggingface.co/datasets/Carol0110/IE-Tagger}{Entropy-Tagger}

%% file: sections/02_introduction.tex
\section{Introduction}
\begin{figure}[t]
    \centering
    \includegraphics[width=\linewidth]{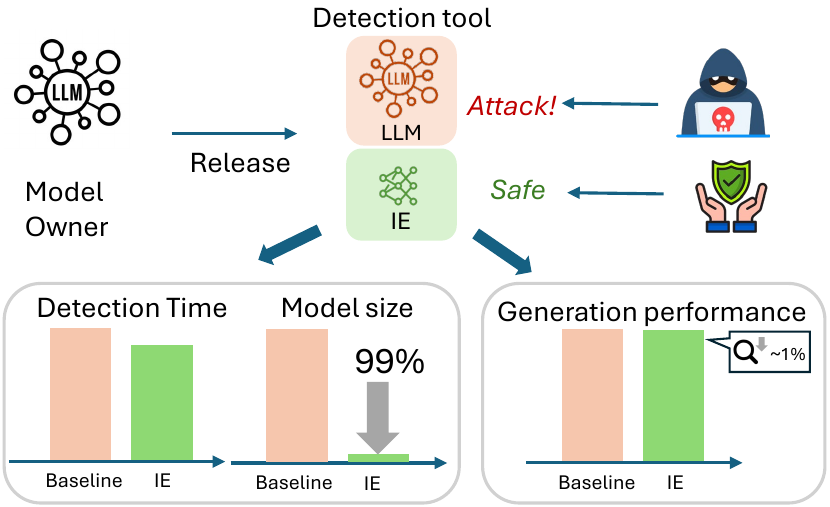}
    \caption{Existing watermarking methods in low-entropy scenarios face safety and cost challenges, while our method addresses them efficiently and~securely.}
    \label{fig:comparison}
\end{figure}
Textual watermarking, which aims to embed subtle patterns in the generated text to make it detectable by algorithms but invisible to humans, is an important step towards trustworthy AI. 
It can be applied at various stages, including logits generation~\cite{kirchenbauer2023watermark}, token sampling~\cite{christ2024undetectable}, and training~\cite{sun2022coprotector,sun2023codemark,gu2024on}. 
Logit-based watermarking is cost-efficient, modifying probabilities before token selection without adding training or sampling steps~\cite{liu2024survey}. 

As a pioneering work in logit-based watermarking, ~\citet{kirchenbauer2023watermark} introduced KGW, the first logit-based watermarking approach. 
This method partitions the vocabulary into green and red lists based on the previous token and a hash key, then boosts the logits of the green list to embed the watermark and decreases the probabilities of tokens outside this green list (red list).
However, KGW fails in \textit{\textbf{low-entropy scenarios}} where the next token is highly predictable, such as the prompt ``import numpy as'' almost certainly leading to``np'' (entropy 0.048).
 If this expected token is placed in the red list, two issues may arise: (1) If the model still selects it despite the reduced probability, the unexpected inclusion of a red-list token may weaken the watermark’s detectability.
(2) If the model instead picks a green-list token due to the boosted logits, it may disrupt text fluency. 
Similarly, if the expected token is directly placed in the green list, it may lead to a \emph{false inflation} of green-list token frequency in generated text, thereby increasing the likelihood of misclassify human-written content as machine-generated. As a result, the watermark detection system becomes less reliable.
 
To address the low-entropy problem, ~\citet{lee-etal-2024-wrote} proposed SWEET, which applies watermarks only to high-entropy tokens, preserving text quality. 
Similarly, ~\citet{lu-etal-2024-entropy} introduced EWD, which enhances detection by assigning higher weights to high-entropy tokens. 
However, these entropy-based watermarking methods face a critical limitation: \emph{they assume the detector has access to the original LLM to calculate entropy.}
This reliance on the original model introduces several challenges, as illustrated in Fig.~\ref{fig:comparison}. 
First, providing the original model to third parties poses significant risks of model leakage, potentially leading to unintended exposure or unauthorized access~\cite{song2020information,duc2014unifying}. 
Second, using the original LLM incurs substantial computational costs, particularly when processing large-scale datasets or running multiple detections. 

Using a proxy model to approximate entropy calculation is potentially feasible. 
SWEET replaces the original model, e.g., LLaMA2-13B~\cite{touvron2023llama}, with a smaller model from the same family, e.g., LLaMA2-7B, for entropy estimation. Although this practice outperforms KGW, it still suffers from significant performance degradation. 
While the original EWD work does not explicitly explore the use of proxy models, our experimental results in Tab.~\ref{tab:main} show a similar trend. It is also important to note that the effectiveness of a proxy model heavily depends on its architectural similarity to the original model.

Motivated by this, we attempt to train a lightweight proxy model to eliminate the dependency on the original LLM during entropy-based watermark detection. Our experiments in App.~\ref{app:training_details_entropy_tagger} show that regressing continuous entropy using an MLP is challenging, but reframing the task as a classification problem -- determining whether the entropy of next token exceeds a given threshold -- is more feasible.
Considering the aforementioned issue that proxy models rely on architectural similarity to the original model, we introduce a Unified Feature Extractor that converts prefix tokens into a unified feature representation using a token translator and an embedding model, thereby ensuring compatibility across different LLMs and tokenizers.
When using a fixed threshold to distinguish high- and low-entropy tokens, we observe that applying the same threshold across all samples ignores inter-sample variability. Moreover, in practical scenarios, the generator and the detector cannot share threshold, which limits the applicability of watermarking methods. To balance the naturalness of generated text and watermark detectability, we propose a sample-level entropy threshold optimization method.
We evaluate our method in a representative low-entropy setting, namely the code generation task, with two widely used benchmarks: HumanEval~\cite{chen2021evaluating} and MBPP~\cite{austin2021program}.

Our main contributions are as follows:
(1) We propose IE, a novel watermarking framework that relies on a small MLP instead of the original LLM to enable safe, efficient and accurate watermark detection.
(2) We present \textit{Threshold Navigator}, a low-high entropy threshold auto-optimization method that enhances detection performance not only for our framework but also for various watermarking approaches.
(3) Our proposed watermarking framework, IE, which integrates the three components, achieves a 99\% reduction in parameter usage while delivering state-of-the-art detection performance, showcasing its efficiency and scalability.

%% file: sections/03_related_work.tex
\section{Related Work}
\noindent\textbf{Traditional Text Watermarking}
 typically modifies generated text to embed watermarks. 
 Based on the granularity of these modifications, existing approaches can be categorized as format-based, lexical-based, syntactic-based, and generation-based methods.
\textit{Format-based} watermarking~\cite{10.1145/2938503.2938510,464718,POR20121075,sato2023embarrassinglysimpletextwatermarks} originates from image watermarking and focuses on altering the text format rather than its content, such as by adjusting text layout or using Unicode-based substitutions.
\textit{Lexical-based} watermarking~\cite{munyer2024deeptextmarkdeeplearningdriventext,pmlr-v202-ni23b,yang2023watermarkingtextgeneratedblackbox,yoo-etal-2023-robust,Yang_Zhang_Chen_Zhang_Ma_Wang_Yu_2022} replaces selected words with their synonyms while preserving the original sentence's syntactic structure. However, this approach is susceptible to attacks involving random synonym replacements.
To address this vulnerability, syntactic-based methods~\citep{10.1007/3-540-45496-9_14,10.1145/1178766.1178777,MERAL2009107} embed watermarks by modifying the text's syntactic structure, which enhances resistance to removal attacks. Nevertheless, these methods often produce unnatural transformations, degrading the quality of the generated text and increasing its susceptibility to detection and targeted attacks.

\noindent\textbf{LLM-Based Watermarking} embeds watermarks in LLMs by intervening at different generation stages, including logits generation, token sampling, and training. 
Watermarking during \textit{logits generation} adjusts the probability distribution over tokens to embed identifiable patterns, while \textit{token sampling}~\cite{christ2024undetectable,kuditipudi2024robust,hou-etal-2024-semstamp,hou-etal-2024-k} modifies the token selection process to incorporate watermarks. 
Watermarks can also be embedded into model weights \textit{during training}~\cite{sun2022coprotector,sun2023codemark,gu2024on,xu2024learning,xu2024hufu}, encoding watermarks into the model itself to ensure traceability and resilience against removal or tampering.

Watermarking during logits generation is the most cost-effective approach, avoiding the overhead of retraining or complex dynamic sampling while remaining flexible for post hoc application. \citet{kirchenbauer2023watermark} proposed the classic vocabulary partitioning method, dividing tokens into "green" and "red" sets, biasing generation toward "green" tokens. Building on this, studies \citep{fernandez2023three,lu-etal-2024-entropy,kirchenbauer2024on} improved detectability, while others \citep{hu2024unbiased,wu2023dipmark,fu2024watermarking,guan-etal-2024-codeip,lee-etal-2024-wrote,chen-etal-2024-watme,liu2024adaptive,wang2024towards,wouters2024optimizing,wang2025morphmarkflexibleadaptivewatermarking} focused on preserving text quality.

To handle low-entropy scenarios, \citet{lee-etal-2024-wrote} focused on watermarking only high-entropy tokens, while \citet{lu-etal-2024-entropy} applied entropy-weighted adjustments to detection statistics. 
However, both approaches rely on re-querying original LLM during detection. 
Our proposed method, IE, eliminates the need for the original LLM during detection, enhancing safety and efficiency.

%% file: sections/04_preliminaries.tex
\begin{figure*}[t]
    \centering    \includegraphics[width=\linewidth]{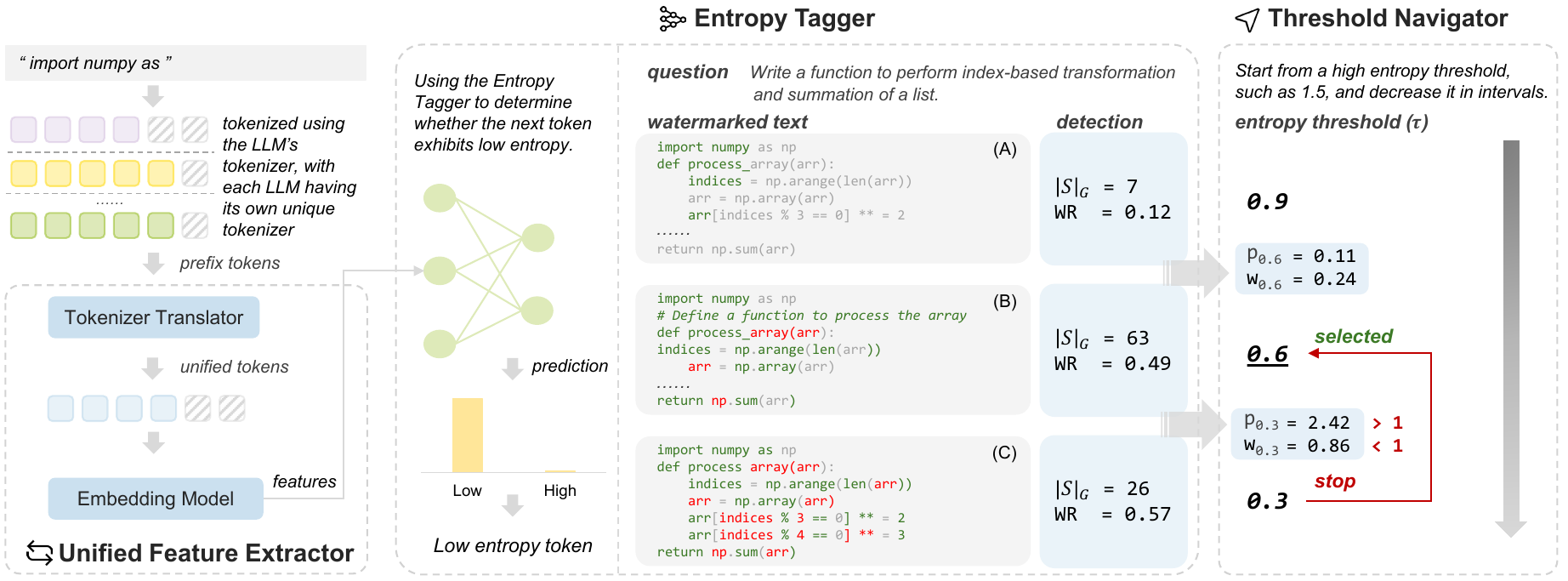}
    \caption{
   \textbf{Overview of IE (Invisible Entropy).}  
The model includes three components: the \textit{Unified Feature Extractor} for tokenizer compatibility and feature extraction, the \textit{Entropy Tagger }to predict if the next token’s entropy exceeds threshold \(\tau\), and the \textit{Threshold Navigator} to optimize \(\tau\) for effective watermarking, naturalness, and robustness. 
Tokens are color-coded as \textcolor[HTML]{FF0000}{red} (red list), \textcolor[HTML]{3B7D23}{green} (green list), and \textcolor[HTML]{7F7F7F}{gray} (unwatermarked).
This example shows the search stopping at \(\tau = 0.6\). At \(\tau = 0.9\), insufficient watermarking occurs, while at \(\tau = 0.3\), excessive low-entropy classification causes token generation issues (e.g., the underscore ``\_'').
    }
    \label{fig:overflow}
\end{figure*}

\section{Preliminaries}
\label{sec:preliminaries}
Our method builds upon the KGW watermarking strategy \cite{kirchenbauer2023watermark} for logits generation. 
KGW operates in two phases: the generation phase and the detection phase.

During the generation phase, when generating the $t$-th token $s_{t}$, a hash key is derived from the previous token $s_{t-1}$.
Using this hash key, the vocabulary is divided into a green list and a red list, with the proportion of green tokens determined by $\gamma$. 
A bias $\delta$ is then added to the logits of tokens in the green list, increasing their likelihood of being selected during sampling.

In the detection phase, for a generated sequence $\{s_1, s_2, \dots, s_{|T|}\}$, where $|T|$ is the number of tokens, the count of green tokens is denoted as $|S|_G$.
A watermark detection statistic $z$ is calculated as:
\begin{equation} \label{eq:z}
\small z = \frac{|S|_G - \gamma |T|}{\sqrt{|T|\gamma(1 - \gamma)}}.
\end{equation}
A detection threshold $\hat{z}$ is predefined. 
If $z > \hat{z}$, the text is classified as watermarked; otherwise, it is considered human-generated. 

%% file: sections/05_methods.tex
\section{Methodology}
In this section, we introduce our IE model in detail. 
The model consists of three modules: the \textit{Unified Feature Extractor}, \textit{Entropy Tagger}, and \textit{Threshold Navigator}, as illustrated in Fig.~\ref{fig:overflow}.

\subsection{Unified Feature Extractor}
Existing works \cite{lee-etal-2024-wrote, lu-etal-2024-entropy} rely on the original LLM to compute exact entropy for determining whether a token has low entropy. However, this approach significantly increases computational costs and the risk of model leakage. In practical applications, knowing the exact entropy value is often unnecessary—binary classification (low or high entropy) is sufficient. Thus, we propose using a smaller model to perform binary entropy prediction.
In this subsection, we introduce a Unified Feature Extractor that learns vector representations of the generated text so far. 
In the next subsection, we present the binary entropy tagger.

Concretely, assume that a sequence of tokens \(\{s_0, s_1, ..., s_{t-1}\}\) has already been generated, and the model is currently generating token \(s_t\). 
These tokens may originate from different tokenizers associated with various LLMs. 
To handle this, our approach employs a \textit{tokenizer translator} that converts prefix tokens into a unified format.
The tokenizer translator first converts the prefix tokens back into raw text and then re-encodes them using the tokenizer of an embedding model. 
The \textit{embedding model} processes the translated tokens and generates unified token embeddings.
While LLMs typically support long input sequences, embedding models—often smaller, encoder-only architectures—are limited by a maximum input length.
To address this, the embedding model focuses on processing only the last segment of tokens, up to its maximum allowable length, ensuring that critical information is retained.
The representation of the last token, \(v_t\), is used to represent the entire generated sequence. This token encapsulates step-by-step contextual dependencies, providing an effective summary of the preceding text for the binary prediction task.

\subsection{Entropy Tagger}
Following the motivation outlined in the previous section, we propose an Entropy Tagger that predicts whether a token \(s_t\) is low-entropy by leveraging the feature vector \(v_t\) obtained from the feature extractor.
The tagger outputs the probability \(p_t\) that the token’s entropy is below a threshold \(\tau\), optimized by binary cross-entropy loss: 
\begin{equation}
\small
\mathcal{L} = -\textstyle \frac{1}{N}\sum_{t=1}^{N} \left[y_t \log(p_t) + (1-y_t) \log(1-p_t)\right],   \nonumber
\label{eq:bceloss}
\end{equation}
where $y_t$ denotes the truth label for the $t$-th sample, computed by the original LLM (0 for high-entropy tokens and 1 for low-entropy tokens), and $N$ is the total number of samples.

For the tagger implementation, we find that a small learnable multi-layer perceptron is sufficient to make accurate predictions. 
For entropy calculation, we employ Shannon entropy \cite{lee-etal-2024-wrote} over the dense Spike Entropy \cite{kirchenbauer2023watermark}, as its dispersed distribution offers clearer boundaries.

\subsection{Threshold Navigator}

The entropy threshold $\tau$  is crucial in balancing watermarked and non-watermarked tokens, directly impacting watermark effectiveness.   When $\tau$  is too high, more tokens are classified as low-entropy, reducing the number of tokens eligible for watermarking, as seen in Block A of Fig. \ref{fig:overflow}, where gray (unwatermarked) tokens dominate.
Conversely, if $\tau$ is too low, fewer tokens are treated as low-entropy, leading to excessive watermarking (e.g., colored tokens (watermarked) dominate in Block C of Fig. \ref{fig:overflow}).
Existing entropy-based watermarking methods rely on manually predefined or empirically determined entropy thresholds~\cite{lee-etal-2024-wrote}, making them less robust since they overlook sample variations and depend heavily on the chosen parameter.  

To address these limitations, we propose our Threshold Navigator. 
The Threshold Navigator automatically searches for an appropriate entropy threshold for each sentence. 
Here, we define an optimistic threshold \(\tau\) as the point\emph{ where the watermark ratio~($\mathrm{WR}$, defined as the ratio of watermarked tokens to the total number of generated tokens) drops while the count of green tokens rises}. 
Intuitively,  a lower watermark ratio indicates lighter modifications to the original text, thereby reducing interference from the watermarking mechanism. 
Meanwhile, an increased count of green tokens signifies better alignment with machine-generated text, making it easier for the watermark to be detected.
We also provide a theoretical proof on this in §\ref{sec:threshold_navigator}.

Based on the above sensitivity analysis, we introduce two metrics. Watermark Ratio Change (\(w\)) measures the change in watermark ratios between entropy thresholds \(\tau_{i-1}\) and \(\tau_i\): \(w_{\tau_i} = \mathrm{WR}_{\tau_{i-1}}/\mathrm{WR}_{\tau_i}\). Green Token Counts Change (\(p\)) quantifies the variation in green token counts: \(p_{\tau_i} = |S|_{G_{\tau_{i-1}}}/|S|_{G_{\tau_i}}\).
During the dynamic adjustment process, the Threshold Navigator lowers the entropy threshold and monitors changes in $\mathrm{WR}$ and the number of green tokens $|S|_G$.
The optimization process stops when \( p > 1 \) and \( w < 1 \), indicating that increasing the entropy threshold improves the sensitivity of the watermarked text to the green token counts while reducing the watermark ratio, thus achieving a balanced and robust distinction. 
Alg.~\ref{alg:threshold_navigator} provides the main procedure.  
An example is shown in Fig.~\ref{fig:overflow}, and additional examples can be found in Fig.~\ref{fig:sample_entropy} in App.~\ref{app:case_study}.  

\input{tables/main_results}

%% file: tables/main_results.tex
\begin{table*}[t]
\centering
\scalebox{0.65}{
\begin{tabular}{clllllllllllll}
\toprule
\multicolumn{2}{c}{\textbf{Method}}& & \multicolumn{5}{c}{\textsc{HumanEval}}  & & \multicolumn{5}{c}{\textsc{MBPP}} \\ \cmidrule{4-8} \cmidrule{10-14}
\multicolumn{2}{l}{} & \textbf{Params}~$\downarrow$ &\textbf{PPR}~$\uparrow$ & \textbf{UES}~$\uparrow$ & \textbf{Pass@1}~$\uparrow$ & \textbf{AUROC}~$\uparrow$ & \textbf{TPR}~$\uparrow$  & &\textbf{PPR}~$\uparrow$ & \textbf{UES}~$\uparrow$& \textbf{Pass@1}~$\uparrow$ & \textbf{AUROC}~$\uparrow$ &  \textbf{TPR}~$\uparrow$ \\ \cmidrule{1-14}
\multicolumn{1}{l}{\textit{\textbf{Post-hoc}}} &  &  & & & & & && &  \\
\multicolumn{1}{l}{} & \logp &{120M} &{5.513} &0.662 &0.334 &0.533 &0.113 & &{5.373} &0.645 &0.378 &0.525 &0.054   \\
\multicolumn{1}{l}{} & \logr &{120M} &{5.583} &0.670 &0.334&0.553 &0.127 & &{5.373} &0.645 &0.378 &0.527 &0.052 \\
\multicolumn{1}{l}{} & \detectgpt &{\phantom{0}1.1B} &0.613 &0.675 &0.334 &0.533 &0.165 & &{0.619} &0.681&0.378 &0.565 &0.158  \\
\multicolumn{1}{l}{} & \detectb &\phantom{0.0}3B &0.220 &0.660 &0.334 &0.549 &0.092& &0.214 &0.643&0.378 &0.531  &0.040\\
\multicolumn{1}{l}{} & \gptzero  &- &- &0.661 &0.334 &0.521 &0.122 & &-&0.619&0.378 &0.449 &0.026 \\
\multicolumn{1}{l}{} & \openai &- &- &0.643 & 0.334 &0.518 &0.053 & &-&0.634&0.378 &0.500  & 0.036 \\ \cmidrule{1-14}
\multicolumn{2}{l}{\textit{\textbf{Watermark-based}}} &  &  &  &   &  &   &  &  & &  \\
& \kgw &-&- &{0.768} &0.253 & 0.904&0.652& &- &{0.732} & 0.242&{0.930}&{0.718}\\ \cmidrule{2-14}
& \textsc{EWD} &15.5B&0.056 &{0.872} & {0.295} & {0.943} & {0.780}  & & 0.051 &{0.790} & {0.293} & {0.930}&{0.678} \\
& \textsc{EWD} &\phantom{0.0}3B&0.290 &0.871 & 0.295 &0.941 &0.778 & &0.256 &0.767 & 0.293 & 0.916&0.602 \\
& \textsc{EWD} &\phantom{0.0}1B&0.861 &0.861 & 0.295 & 0.931 &0.745  & &0.757 &0.757 & 0.293 &0.910&0.567 \\ \cmidrule{2-14}
& \sweet &15.5B &0.057 &{0.884} &{0.301} &{0.944}   &{0.789}   & &0.051 & {0.785} &{0.322}&{0.901}&{0.536}\\
& \sweet &\phantom{0.0}3B &0.264 &0.792 &0.253 &0.933 &0.722  & &0.245 &0.737 &0.293 &0.896 &0.500\\
& \sweet &\phantom{0.0}1B &0.764 &0.764 & 0.253 &0.925 &0.615  & & 0.732 &0.732 &0.293 & 0.891&0.487\\\cmidrule{2-14}
\multirow{-2}{*}{}&\ie&{130M}&{6.709} &{0.872}  &{0.294} & {0.941}&{0.787} &&{5.805}&{0.755}&{0.301}&{0.892}&0.534\\
\bottomrule
\end{tabular}
}
\caption{\textbf{Main results on \textsc{HumanEval} and \textsc{MBPP}.} "-" indicates either undisclosed parameters~(e.g., GPTZero, OpenAI Classifier) or no additional models required~(e.g., KGW).}
\label{tab:main}
\vspace{-0.2in}
\end{table*}

%% file: sections/06_experiments.tex
\section{Experiments}
\subsection{Tasks and Metrics}
We evaluate IE and baselines in two Python code generation tasks: HumanEval~\cite{chen2021evaluating} and MBPP~\cite{austin2021program}. 
We assess IE and baselines in effectiveness and efficiency.

The evaluation of \textit{effectiveness} focuses on both code generation ability and detectability.
We assess code generation using Pass@$k$, and detectability using AUROC, which measures the model's ability to distinguish watermarked from non-watermarked text. 
We also report the True Positive Rate (TPR), which measures the proportion of correctly identified machine-generated text when the False Positive Rate (FPR) is less than 5\%.  
We propose the Unified Effectiveness Score (UES), averaging the normalized Pass@1 and detectability metrics for overall evaluation: $\text{UES} = \textstyle \textstyle \frac{\frac{\text{Pass@1}}{\mathrm{Pass@1}_{\text{non}}} + \left(\frac{\text{AUROC} + \text{TPR}}{2}\right)}{2}$, where \(\mathrm{Pass@1}_{\text{non}}\) represents the Pass@1 for text without watermark.

From an \textit{efficiency} standpoint, we highlight the number of parameters, denoted as Params, necessary for watermarking in the detection phase.
The detection time required by the watermarking methods is also reported in Tab.~\ref{tab:time}.

To combine effectiveness and efficiency, we introduce a new metric called Performance-to-Params Ratio (PPR), defined as: $\textsc{PPR} = \textstyle \frac{\text{UES}}{\text{Params}}$.

\subsection{Baselines}
\begin{figure*}[t]
    \centering
    \includegraphics[width=0.9\linewidth]{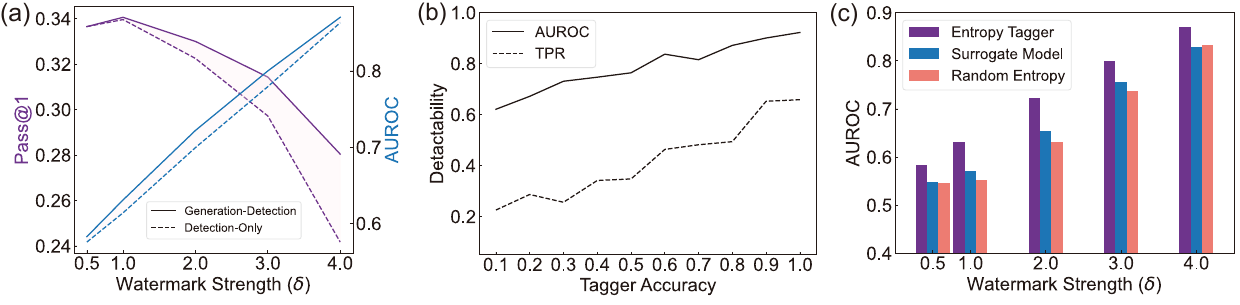}
    \caption{\textbf{Analysis of the Entropy Tagger. }(a) Comparison of applying the Entropy Tagger at different stages: generation-detection versus detection-only. (b) The relationship between Entropy Tagger accuracy and its effectiveness in watermarking. (c) Demonstration of the superior performance of the Entropy Tagger compared to a surrogate model and randomly set entropy.}
    \label{fig:analysis_entropy}
\end{figure*}
We compare IE with post-hoc detection baselines and watermarking methods.
Post-hoc detection does not require any modification during the generation process, thus maintaining the same text quality as non-watermarked text.
LogP(x) and LogRank~\cite{gehrmann-etal-2019-gltr}, and DetectGPT~\cite{pmlr-v202-mitchell23a} are zero-shot detection methods that do not require labeled data. In contrast, GPTZero~and OpenAI Classifier~\cite{solaiman2019releasestrategiessocialimpacts} are trained classifiers.

We select KGW~\cite{kirchenbauer2023watermark}, SWEET~\cite{lee-etal-2024-wrote}, and EWD~\cite{lu-etal-2024-entropy} as watermarking methods for comparison. 
KGW applies watermarking to all tokens during both the generation and detection phases. 
SWEET only applies watermarking to low-entropy tokens during both phases, resulting in higher text quality and detectability compared to KGW (see details in App.~\ref{app:algorithms_for_sweet}).
EWD improves text watermarking detection by assigning higher influence weights to higher-entropy tokens during detection.
To explore the performance SWEET and EWD on smaller surrogate models, we also provide experimental results using StarCoder-3B and StarCoder-1B to compute entropy. In this setting, KGW serves as the watermark generator, while SWEET and EWD act as detectors.

\subsection{Implementation}
In our implementation, we use Starcoder~\cite{li2023starcoder} as the LLM and SimCSE~\cite{gao2021simcse} as the embedding model. 
We use MBPP dataset to train Entropy Tagger, where details are in App.~\ref{app:training_details_entropy_tagger}.
For the post-hoc methods, KGW and SWEET, we adopt the optimal hyperparameters reported by ~\citealp{lee-etal-2024-wrote}. While for EWD, we follow the settings in \citet{lu-etal-2024-entropy}. Since SWEET provides results corresponding to specific entropy threshold, we calculate the average of the results across these different entropy thresholds.
For IE, we use the optimal hyperparameters $\gamma=0.5$ and $\delta=3.0$ unless otherwise specified.
All experiments can be conducted on one single A100-40G. More detailed settings are provided in App.~\ref{app:implementaton_baselines}. 

\subsection{Main Results}
We show the main results in Tab.~\ref{tab:main}.

From \textbf{Effectiveness} perspective, we can draw the following conclusions:
(1)~{\emph{Post-hoc methods fail to handle machine-generated text in low-entropy scenarios.} The UES of all watermark-based methods exceeds 0.75 on the HumanEval dataset and 0.70 on the MBPP dataset, whereas post-hoc methods remain below 0.70 on both datasets.}
(2)~\emph{Our IE demonstrates strong effectiveness,} outperforming post-hoc methods and achieving comparable performance to SWEET and EWD.
(3) \emph{SWEET and EWD suffer performance degradation when applied with smaller models.} When using a surrogate model, IE~(130M) significantly outperforms SWEET~(1B/3B). While EWD is less sensitive to the choice of surrogate model compared to SWEET, it still underperforms IE on HumanEval.
From an \textbf{Efficiency} perspective, LogP(x) and LogRank use BERT with 0.12B parameters for detection. \textsc{DetectGPT} relies on SantaCoder (1.1B) or T5-3B (3B). GPTZero and OpenAI Classifier are closed-source, with parameter counts unavailable. KGW requires no additional model, while SWEET and EWD depend on the original LLM (15.5B).
In contrast, our method uses an embedding model and a lightweight MLP, totaling 0.13B parameters, comparable to Post-hoc methods. 

We finally use PPR to evaluate the \textbf{combined effectiveness and efficiency} of the methods.
Among all methods, IE achieves the highest PPR, significantly outperforming other watermarking methods. 
While Post-hoc methods like LogP(x) and LogRank achieve relatively higher PPRs compared to weaker baselines, their effectiveness remains low.

%% file: sections/07_analysis_and_discussions.tex
\section{Analysis and Discussion}
\subsection{Analysis on Entropy Tagger}
\label{sec:entropy_tagger}
\begin{figure*}[t]
    \centering
    \includegraphics[width=\linewidth]{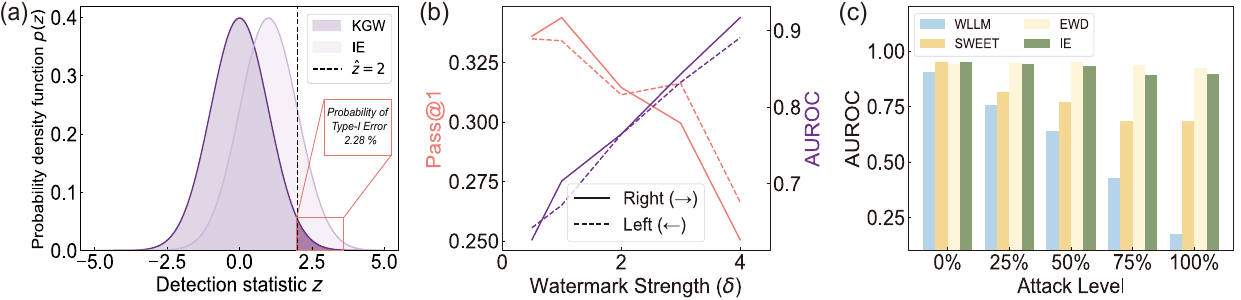}
    \caption{(a) Type-I Error probability and the distribution of detection statistic \(z\) for human-written text. (b) Impact of threshold navigator search directions. (c) Robustness of detection to paraphrasing attacks.}
    \label{fig:analysis_navigator}
\end{figure*}
\noindent\textbf{Generation-Detection or Detection-only?}
We compare the performance of the Entropy Tagger in two setups: Detection-only and Generation-Detection. 
In the Detection-only setup, ground truth entropy values are used to classify tokens as high or low entropy, with a fixed threshold applied for watermark detection. In contrast, the Generation-Detection setup incorporates the Entropy Tagger during the generation phase, predicting entropy values to embed watermarks dynamically.
As shown in Fig.~\ref{fig:analysis_entropy}(a), experimental results indicate that Generation-Detection consistently outperforms Detection-only in both Pass@1 and AUROC across different watermark strengths. This demonstrates that aligning entropy-aware methods during both generation and detection is essential for achieving robust and effective watermarking.

\noindent\textbf{Relationship between Entropy Tagger Accuracy and Watermark Detectability.}  
We investigate how Entropy Tagger accuracy impacts watermark detectability by varying the tagger's accuracy and observing its effect on detection metrics. Under a Detection-only setting, we calculate the exact entropy of watermarked text and simulate tagger inaccuracies by introducing disturbances, where the disturbance proportion \(r\) (0.0 to 1.0) determines the tagger’s accuracy as \(1 - r\).  
Results in Fig.~\ref{fig:analysis_entropy}(b) show that higher tagger accuracy leads to improved AUROC and TPR, highlighting the importance of precise entropy predictors for robust watermarking.  

\noindent\textbf{Comparison with surrogate and random entropy.}
We also replace the Entropy Tagger in the IE framework with a surrogate model~(StarCoder-3B) and with Random Entropy~(a floating-point value randomly selected between -5.0 and 5.0), respectively. As shown in Fig.~\ref{fig:analysis_entropy}(c), the results demonstrate that using the Entropy Tagger significantly outperforms both the Surrogate Model and Random Entropy. Furthermore, the Entropy Tagger contains only 0.13B parameters, making it significantly more cost-effective than the surrogate model. These comparisons confirm the superiority of Entropy Tagger both effectively and efficiently.

\subsection{Analysis on Threshold Navigator}
\label{sec:threshold_navigator}

\noindent\textbf{Theoretical Validation}  
The primary goal of watermark detection is to minimize Type-I and Type-II errors. Thus, we analyze the impact of the Threshold Navigator on both. Generally, our analysis shows that it has no impact on Type-I Error but significantly reduces Type-II Error.  

\textit{\textbf{Type-I Error}} measures the probability of human-written text being misclassified as watermarked. 
For human-written text \( T \), each token is assumed to be independent of the watermarking algorithm, and the probability of a token being included in the green list is denoted by \( \gamma \). As a result, the number of green tokens \( |S|_G \) follows a normal distribution: $|S|_G \sim \mathcal{N}(\gamma |T|, \gamma(1 - \gamma)|T|)$.
In the case of selective watermarking methods such as SWEET, where only a portion of tokens are watermarked, the distribution becomes: $|S|_G \sim \mathcal{N}(\gamma |\tilde{T}|, \gamma(1 - \gamma)|\tilde{T}|)$.
Here, \( |\tilde{T}| = \mathrm{WR} \times |T| \) represents the fraction of the text covered by the watermark. 
Regardless of the value of \( \mathrm{WR} \), the distribution can be standardized using:  $z = \frac{|S|_G - \gamma |\tilde{T}|}{\sqrt{\gamma(1 - \gamma)|\tilde{T}|}}$.
Because \( |S|_G \) follows a normal distribution, the standardized variable \( z \) follows a standard normal distribution \( \mathcal{N}(0, 1) \). 
The probability density function \( p(z) \) describes the likelihood of observing a specific value of \( z \), and the Type-I Error corresponds to the area under the standard normal curve beyond a given threshold (e.g., when \( z = 2 \), the error is 2.28\%, shown as the red region in Fig.~\ref{fig:analysis_navigator}(a)). Since the probability of $z>\hat{z}$ for human text remains constant across $\tau$, the selection of \( \tau \) does not affect the Type-I Error rate.

\textit{\textbf{Type-II Error}} measures the probability of watermarked text being misclassified as human-written text, with lower Type-II Error indicating better detection performance. 
To show how the Threshold Navigator reduces Type-II Error, we analyze its search criterion (\(p > 1\) and \(w < 1\)) and its effect on the detection statistic \(z\), as higher \(z\) values directly lower Type-II Error. 
Specifically, we examine the relationship between \(z\) and two key factors: green token count (\(|S|_G\)) and watermark ratio (\(\mathrm{WR}\)). 
For selective watermarking methods (e.g., IE or SWEET), \(z\) can be expressed as: 
\[
z = \frac{|S|_G - \gamma \cdot \mathrm{WR} \cdot |T|}{\sqrt{\mathrm{WR} \cdot |T| \cdot \gamma (1-\gamma)}}.
\]
A higher $z$ for machine-generated text indicate better watermark detectability, as $z$ quantifies the statistical deviation of the green token count from its expected value in human text.  

To understand how \(z\) changes with \(|S|_G\) and \(\mathrm{WR}\), we compute the partial derivatives of \(z\):  
{\small
\[
\frac{\partial z}{\partial |S|_G} = \frac{1}{\sqrt{\mathrm{WR} \cdot |T| \cdot \gamma (1-\gamma)}} > 0,
\]
}
showing that \(z\) is positively correlated with \(|S|_G\).  
\[
\resizebox{0.48\textwidth}{!}{$
\frac{\partial z}{\partial \mathrm{WR}} = -\frac{|S|_G}{\sqrt{|T| \cdot \gamma (1-\gamma)}} \cdot \frac{1}{2 \cdot \sqrt{\mathrm{WR}^3}} - \sqrt{\frac{\gamma |T|}{1-\gamma}} \cdot \frac{1}{2 \sqrt{\mathrm{WR}}} < 0,
$}
\]
showing that \(z\) is negatively correlated with \(\mathrm{WR}\).   

These results show that increasing \(|S|_G\) improves \(z\) under the condition \(p > 1\), allowing green token counts to grow as thresholds adjust. Simultaneously, decreasing \(\mathrm{WR}\) enhances \(z\) under \(w < 1\), reducing watermarked tokens and improving detectability.
Therefore, our Threshold Navigator effectively reduces Type-II Error by optimizing \( |S|_G \) and \( \mathrm{WR} \), leading to improved watermark detection.

\noindent\textbf{Impact of Search Directions}
\label{subsec:directional}
\begin{figure*}[t]
    \centering    \includegraphics[width=0.9\linewidth]{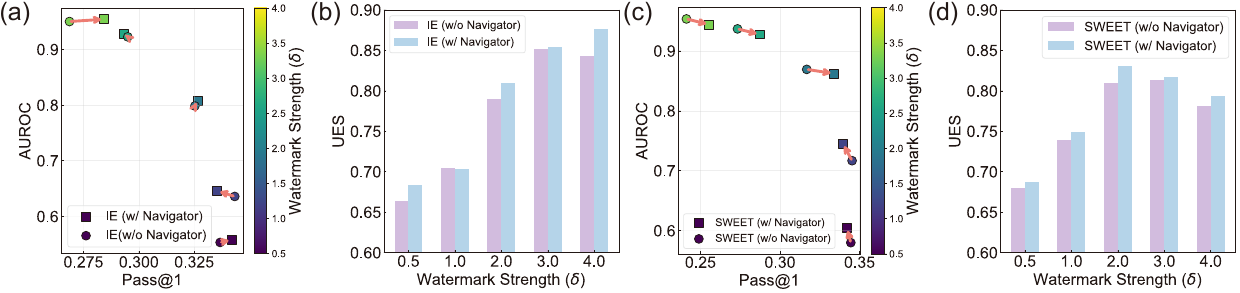}
    \caption{\textbf{Effectiveness of the Threshold Navigator.}
(a) Improved detectability and quality with the Navigator across \(\delta\).  
(b) Improved UES with the Navigator.
(c) Generalizability to SWEET: Pass@1 vs. AUROC, demonstrating similar improvements.  
(d) UES comparison for SWEET, showing significant gains with the Navigator.}
\label{fig:effectiveness}
\end{figure*}
Our default threshold search proceeds from high to low.  
Since different search directions may impact the results, we compare searches starting from high to low ($\leftarrow$) and low to high ($\rightarrow$) to assess their effects.
The experimental results are shown in Fig.~\ref{fig:analysis_navigator}(c). It can be observed that as the watermarking strength increases, navigation towards the Right generally achieves higher AUROC in most cases. 
Conversely, when the watermarking strength is relatively low, navigation towards the Left results in better code quality. 
This is because, at higher watermarking strengths, the impact on code quality becomes more significant, and navigation towards the Left, which prioritizes selecting higher entropy thresholds, helps mitigate the degradation of code quality.

\noindent\textbf{Effectiveness and Orthogonality.}
Fig.~\ref{fig:effectiveness}(a) presents an ablation study where the Threshold Navigator is removed, showing the Pass@1 and AUROC of the watermark under different watermark strengths. 
Fig.~\ref{fig:effectiveness}(b) illustrates the UES across the same range of watermark strengths. 
These results demonstrate that the Threshold Navigator significantly enhances the AUROC  and UES of IE, enabling the output to strike a balance between quality and detectability. 
To further evaluate the generalizability of the Threshold Navigator across different watermark backbones, we apply it to the SWEET watermarking method. 
As shown in Fig.~\ref{fig:effectiveness}(c,d), the Navigator significantly improves SWEET across various watermark strengths. 
This highlights the versatility of the Threshold Navigator, as it can be seamlessly integrated with existing watermarking methods to enhance their effectiveness. 

\subsection{Robustness to Paraphrasing Attacks}
Malicious users may attempt to remove the watermark by paraphrasing attacks~\cite{krishna2023paraphrasing,gao2024shaping}. 
Here, we conduct variable name paraphrasing attacks on the generated codes at different levels. 
Specifically, for the generated codes from each watermarking method on the HumanEval dataset, we replace varying proportions of variable names in the watermarked text. 
Fig.~\ref{fig:analysis_navigator}(c) shows the detectability of the attacked code, measured by AUROC. 
The Attack Level denotes the percentage of variable names changed, with 0\% meaning none are altered and 25\% indicating a quarter are paraphrased.
It can be seen that as the Attack Level increases, the detectability of all methods declines. 
Notably, KGW and SWEET experience the most significant drops, with KGW’s detectability falling below 20\% and SWEET dropping below 80\%.
Meanwhile, IE and EWD show better robustness, maintaining around 90\%. 

%% file: sections/08_conclusion.tex
\section{Conclusion}
We introduce IE (Invisible Entropy), a selective watermarking method that overcomes two key limitations: reliance on the original LLM for costly entropy calculations and difficulty watermarking predictable, low-entropy outputs.
IE uses a lightweight feature extractor and entropy tagger to predict token entropy without the original LLM and a Threshold Navigator for adaptive entropy thresholds, ensuring balance in effectiveness, naturalness, and detectability. 
Experiments on HumanEval and MBPP show a 99\% parameter reduction with state-of-the-art performance.
In the future, we aim to further enhance the accuracy of the entropy tagger to improve watermarking effectiveness and robustness.

%% file: sections/09_limitations.tex
\section*{Limitations}
Although IE offers a safe, efficient and accurate watermarking approach, we identify two limitations and suggest potential solutions to address them.

\paragraph{Entropy Tagger Accuracy Calibration}
In the App.~\ref{app:training_details_entropy_tagger}, we report the accuracy of the trained Entropy Tagger. Although the current Entropy Tagger performs comparably to the precise entropy calculation, there is still some slight decrease in performance. Therefore, future work could focus on training a more precise Entropy Tagger, such as by incorporating certain specific low-entropy tokens as analyzed in App.~\ref{app:analysis_low_entropy}.

\paragraph{Optimization Strategy for Threshold Navigator}
In \S~\ref{subsec:directional}, we analyze the impact of the two search directions of the Threshold Navigator on watermarking performance.
However, in our experiments, the search granularity is fixed at 0.3, which may limit optimization flexibility. 
Future work could explore adaptive search granularities that dynamically adjust based on context or performance feedback, as well as alternative search directions that better align with different watermarking scenarios to further enhance performance.

%% file: appendicies/sample_for_navigator.tex
\section{Case study for threshold navigator}
\label{app:case_study}
\begin{figure}[!hbtp]
    \centering
    \begin{minipage}{0.45\textwidth}
        \centering
        \includegraphics[width=\linewidth]{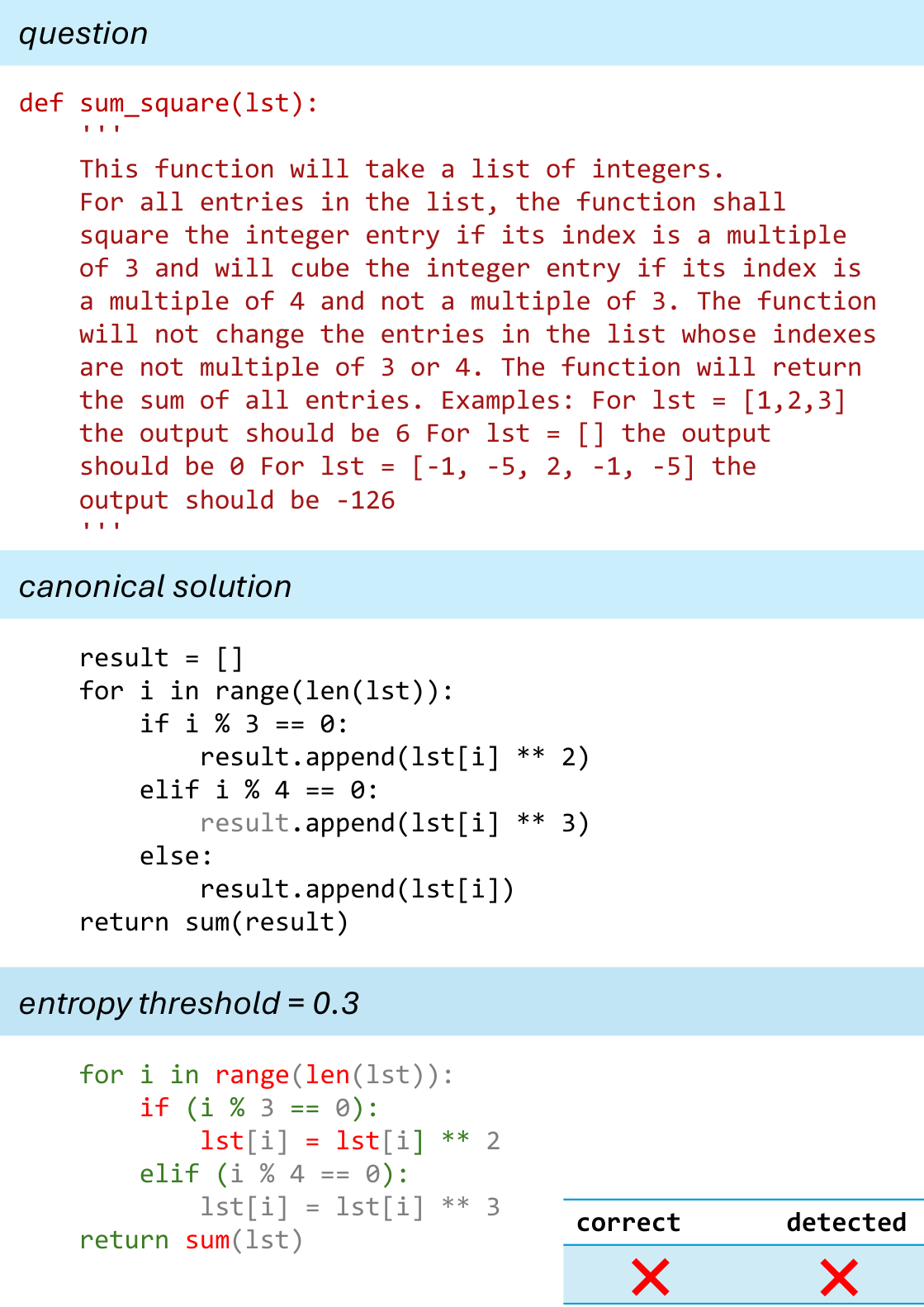}
    \end{minipage}
    \hfill
    \begin{minipage}{0.45\textwidth}
        \centering
        \includegraphics[width=\linewidth]{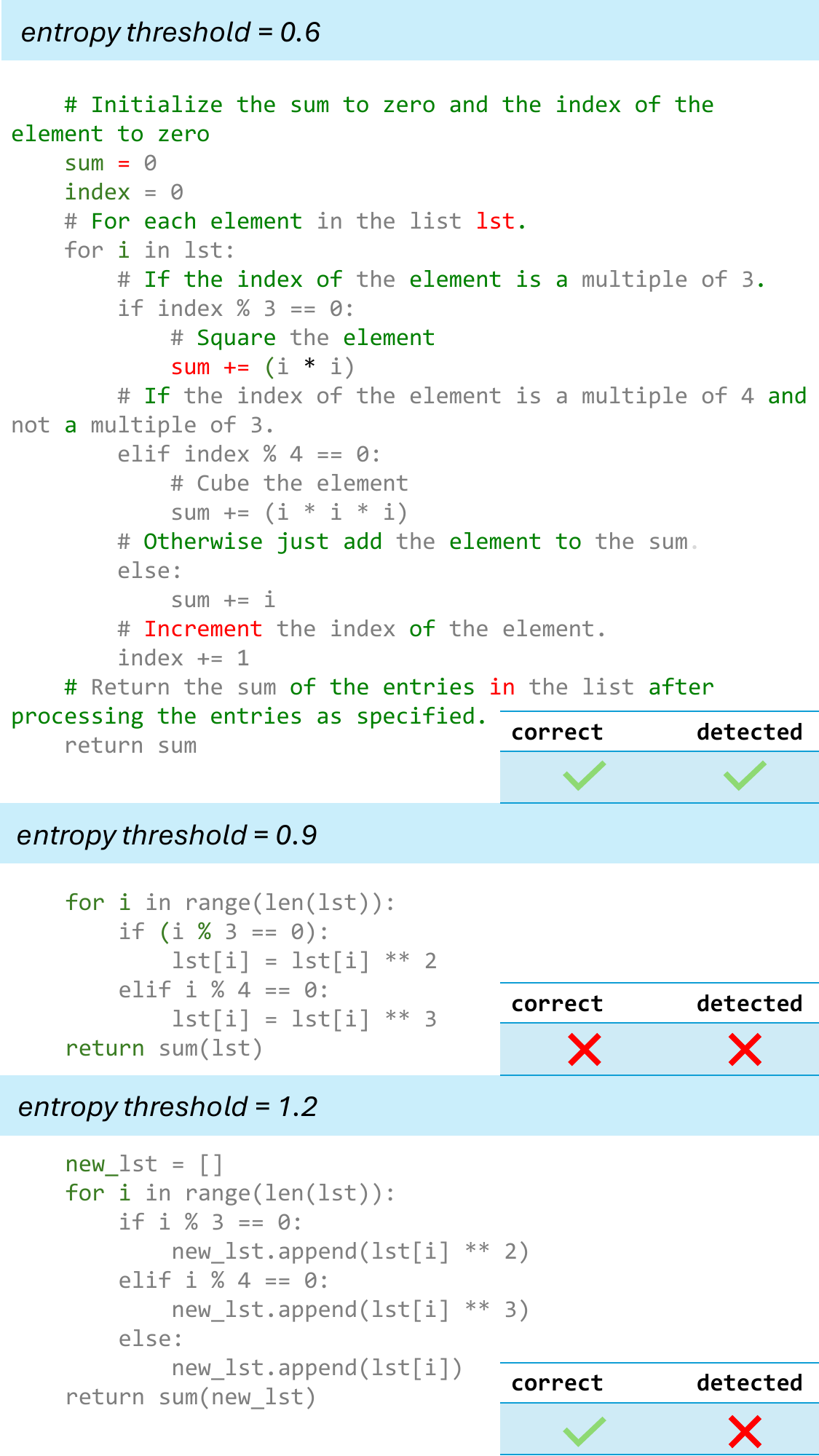}
    \end{minipage}
    \caption{Results for various entropy thresholds.}
    \label{fig:sample_entropy}
\end{figure}

In this section, we present a case study on Threshold Navigator. We select different entropy thresholds $\tau$~(0.3, 0.6, 0.9, and 1.2), where token below the entropy threshold are not watermarked.
The experimental results are shown in Fig.~\ref{fig:sample_entropy}. In the watermarked text, tokens are annotated in \textcolor[HTML]{FF0000}{red}, \textcolor[HTML]{3B7D23}{green}, or \textcolor[HTML]{7F7F7F}{gray} to represent red tokens, green tokens, and unwatermarked tokens, respectively. The evaluation is conducted from two perspectives, correct~(whether the code correctly answer the question) and detected~(whether the watermark is successfully detected).

It is indicated that when $\tau$ is set to 0.3, the proportion of watermarked token is relatively high, which tends to result in lower code correctness. Conversely, when $\tau$ is set to 1.2, the proportion of watermarked tokens is relatively low. While this helps maintain code correctness to some extent, it also leads to a decrease in watermark detectability.
Using the Threshold Navigator algorithm, the results are shown in Tab.~\ref{tab:p_w_case}. When $\tau$ is set to 0.3, the values of $p$ and $w$ satisfy the condition $p>1$ and $w<1$, respectively. Therefore, a "transition" is required for 0.3, leading to the correct selection of 0.6 as entropy threshold. This is further validated in Fig.~\ref{fig:sample_entropy}, where an entropy threshold of 0.6 ensures both correctness and detectability.

\begin{table}[!htbp]
\centering
\small
\begin{tabular}{llll}
\toprule
Entropy Threshold                  & 0.3  & 0.6  & 0.9  \\ \midrule
$p$                                & 3.57 & 0.12 & 0.75 \\ \midrule
$w$                                & 0.98 & 0.29 & 0.58 \\ \midrule
$p > 1$ and $w < 1$?                & Yes    & No    & No    \\ \bottomrule
\end{tabular}
\caption{$p$ and $w$ under different entropy thresholds.}
\label{tab:p_w_case}
\end{table}

%% file: appendicies/algorithms_for_watermark_generation_and_detection.tex
\section{Algorithms for entropy-based selective watermark (SWEET)}
\label{app:algorithms_for_sweet}
In this section, we present the algorithms for entropy-based selective watermark generation and detection (SWEET), as shown in Alg.~\ref{alg:text_generation_sweet} and Alg.~\ref{alg:text_detection_sweet}. The core idea has already been introduced in the \S ~\ref{sec:preliminaries}, while watermarking is applied only to the tokens with entropy greater than $\tau$ during generation and detection process.

The algorithm for text generation with entropy-based selective watermarking is built on KGW, as shown in Alg.~\ref{alg:text_generation_sweet}. Initially, the language model processes the preceding tokens to compute the probability distribution \( p^{(t)} \) over the vocabulary for the next token \( s_t \) (Line 3). The entropy of this distribution determines whether the watermark is applied (Line 4). If the entropy \( H_t \) exceeds a threshold \( \tau \), the vocabulary is partitioned into a "green list" and a "red list" using a hash function seeded by the previous token. The size of the green list is controlled by a proportion parameter \( \gamma \), and its logits are increased by a hardness parameter \( \delta \) to influence token selection. The final token is sampled from the adjusted probability distribution (Lines 5 to 9). If the entropy \( H_t \) is below the threshold, the token is sampled from the original distribution without modification (Line 11).

The detection phase for entropy-based selective watermarking is similar to the generation phase, as shown in Alg.~\ref{alg:text_detection_sweet}. It initializes counters for green list tokens (\( |S|_G \)), scored tokens (\( |\hat{T}| \)), total generated tokens (\( |T| \)), and the Watermark Ratio (\( \text{WR} \)) (Line 2). For each token, the entropy \( H_t \) is computed (Line 4). If \( H_t \) exceeds the threshold \( \tau \), a hash of the previous token seeds a random number generator to partition the vocabulary into a green list \( G \) and a red list \( R \). Tokens in the green list increment the green token count, while all scored tokens update the scored token count (Lines 5 to 9). After processing all tokens, a standardized score \( z \) is calculated to measure the deviation in green token frequency and the Watermark Ratio \( \text{WR} \) (Line 13). If \( z \) exceeds a predefined threshold \( \hat{z} \), the text is classified as watermarked; otherwise, it is considered unwatermarked (Lines 14 to 18). 

\begin{algorithm}[htbp]
\caption{Text Generation with entropy-based selective watermark}
\label{alg:text_generation_sweet}
\begin{algorithmic}[1]
    \STATE \textbf{Input:} \parbox[t]{\dimexpr\linewidth-\algorithmicindent}{%
        prompt, $s_{-N_p}, \ldots, s_{-1}$ \\
        entropy threshold, $\tau$ \\
        green list size, $\gamma\in(0,1)$ \\
        hardness parameter, $\delta>0$
    }
    \FOR{$t=0,1,...$}
        \STATE Apply the language model to prior tokens $s_{-N_p}, \ldots, s_{-1}$ to get a probability vector $p^{(t)}$ over the vocabulary.
        \STATE Calculate the entropy $H_{t}$ for next token $s_{t}$.
        \IF{$H_{t} > \tau$}
            \STATE Compute a hash of token $s_{t-1}$, and use it to seed a random number generator.
            \STATE Using this random number generator, randomly partition the vocabulary into a "green list" $G$ of size $\gamma|V|$, and a "red list" $R$ of size $(1-\gamma)|V|$.
            \STATE Add $\delta$ to each green list logit. Apply the softmax operator to these modified logits to get a probability distribution over the vocabulary. \\
            \begin{equation*}
            \hat{p}_{k}^{(t)}=\left\{
            \begin{array}{ll}
            \frac{e^{ \left( l_{k}^{(t)} + \delta \right) }}{\sum_{i \in R} e^{ l_{i}^{(t)} } + \sum_{i \in G} e^{ l_{i}^{(t)} + \delta }}, & k \in G \\
            \frac{e^{ l_{k}^{(t)} }}{\sum_{i \in R} e^{ l_{i}^{(t)} } + \sum_{i \in G} e^{ l_{i}^{(t)} + \delta }}, & k \in R
            \end{array}
            \right.
            \end{equation*}
            \STATE Sample the next token, $s_{t}$, using the marked distribution $\hat{p}^{(t)}$.
        \ELSE
            \STATE Sample the next token, $s_{t}$, using the origin distribution $p^{(t)}$. 
        \ENDIF
        \ENDFOR
\end{algorithmic}
\end{algorithm}

\begin{table}[p]
\centering
\scalebox{0.85}{
\begin{tabular}{llll}
\toprule
\textbf{Dataset} & \textbf{Split} &\textbf{\# Samples}&\textbf{\# Converted} \\ \midrule
HumanEval    & test       & 164 & 32,168     \\ \midrule
\multirow{3}{*}{MBPP}         & train      & 374 & 29,747        \\
             & validation & \phantom{0}90  &  \phantom{0}7,391     \\
             & test       & 500 & 40,571       \\ 
 \bottomrule
\end{tabular}
}
\caption{\textbf{Statistics of HumanEval and MBPP.} \#Samples indicates the number of samples in each split of the dataset, while \#Converted represents the number of samples in each split after preprocessing.}
\label{tab:statistics_humaneval_mbpp}
\end{table}

\begin{algorithm}[p]
\caption{Detection with entropy-based selective watermark}
\label{alg:text_detection_sweet}
\begin{algorithmic}[1]
    \STATE \textbf{Input:} \parbox[t]{\dimexpr\linewidth-\algorithmicindent}{
        prompt, $s_{-N_p}, \ldots, s_{-1}$ \\
        entropy threshold, $\tau$ \\
        green list size, $\gamma\in(0,1)$ \\
        z threshold, $\hat{z}$
    }
    \STATE \textbf{Initialize:} \parbox[t]{\dimexpr\linewidth-\algorithmicindent} {
        green token counts, $|S|_G \gets 0$ \\
        scored tokens counts, $|\hat{T}| \gets 0$ \\
        generated tokens counts, $|T| \gets 0$ \\
        watermark ratio, $\text{WR} \gets 0$
    }
    \FOR{$t=0,1,...$}
        \STATE Compute the entropy $H_t$ of the next token $s_{t}$.
        \IF{$H_t>\tau$}
            \STATE Compute a hash of $s_{t-1}$, and use it to seed a random number generator.
            \STATE Using the random number generator, randomly partition the vocabulary into a "green list" $G$ of size $\gamma|V|$, and a "red list" $R$ of size $(1-\gamma)|V|$.
            \STATE Increment $|S|_G$ if $s_{t}$ in green list.
                \[
                |S|_G \gets \left\{
                \begin{array}{ll}
                |S|_G + 1, & \text{if } s_t \in G \\
                |S|_G, & \text{otherwise}
                \end{array}
                \right.
                \]
            \STATE Increment $|\hat{T}| \gets |\hat{T}|+1$
        \ENDIF
        \STATE Increment $|T| \gets |T|+1$
    \ENDFOR
    \STATE Compute $z$ and $\text{WR}$.
    \begin{align*}
        z &= \frac{|S|_G - \gamma |\hat{T}|}{\sqrt{\gamma(1-\gamma)|\hat{T}|}}, \\
        \mathrm{WR} &= \frac{|\hat{T}|}{|T|}
    \end{align*} \noindent
    \STATE return $z > \hat{z}, \text{WR}, |S|_G$
\end{algorithmic}
\end{algorithm}

\begin{table}[p]
\centering
\scalebox{0.85}{
\begin{tabular}{c|c|c|c}
\toprule
\textbf{Entropy} & \textbf{MBPP} &\textbf{MBPP} & \textbf{HumanEval} \\ 
\textbf{Threshold}& \textbf{Validation} & \textbf{Test} & \textbf{Test} \\ \midrule
0.3               & 83.89           & 81.93     & 68.47          \\
0.6               & 82.52           & 81.06     & 66.61          \\
0.9               & 83.45           & 82.31     & 68.51          \\
1.2               & 84.54           & 83.73     & 70.95          \\
1.5               & 86.97           & 86.79     & 75.71          \\ \bottomrule
\end{tabular}
}
\caption{Accuracy of Entropy Tagger for different entropy thresholds.}
\label{tab:accuracy_entropy_tagger}
\end{table}

%% file: appendicies/training_details_of_entropy_tagger.tex
\section{Training details of entropy tagger}
\label{app:training_details_entropy_tagger}
\subsection{Preprocess}
The statistics of HumanEval and MBPP datasets is shown in Tab.~\ref{tab:statistics_humaneval_mbpp}. During the preprocessing phase, we use the training split of the MBPP dataset to construct the training dataset for the Entropy Tagger, with the preprocessing algorithm described in Alg.~\ref{alg:preprocess}. Specifically, we first concatenate the prompt with the code.~(Line 4) Next, we truncate the sequence starting from the beginning, adding one token at a time, and compute the exact entropy using LLM as the label. Then, we use the Unified Feature Extractor to extract features from the truncated sequence to obtain the feature vector $v$.~(Lines 5 to 12) Finally, we obtain the preprocessed dataset $\hat{D}=\{(X_i,y_i)\}$, where $X_i$ represents the $i$-th feature vector $v$, and $y_i$ represents the corresponding actual entropy for $X_i$. The dataset size for each split is shown in~Tab.~\ref{tab:statistics_humaneval_mbpp}.

\begin{algorithm}[t]
\caption{Algorithm for preprocessing of Entropy Tagger}
\label{alg:preprocess}
\begin{algorithmic}[1]
    \STATE \textbf{Input:} Original dataset $D = \{T_i\}$, where $T_i$ represents a sample containing a prompt and corresponding code.
    \STATE \textbf{Output:} Preprocessed dataset $\hat{D} = \{(X_i, y_i)\}$, where $X_i$ is the feature vector and $y_i$ is the actual entropy.
    
    \FOR{each sample $T_i$ in $D$}
        \STATE Concatenate the prompt and code in $T_i$ to form a single sequence $S$.
        \STATE Initialize an empty list $\hat{S} = []$ to store truncated sequences.
        \FOR{$k = 1$ to $\text{length}(S)$}
            \STATE Truncate $S$ to the first $k$ tokens to create $S_k$.
            \STATE Append $S_k$ to $\hat{S}$.
        \ENDFOR
        \FOR{each truncated sequence $S_k$ in $\hat{S}$}
            \STATE Compute the exact entropy $y_k$ of $S_k$ using StarCoder.
            \STATE Extract the feature vector $v_k$ for $S_k$ using the Unified Feature Extractor.
            \STATE Add $(v_k, y_k)$ to $\hat{D}$.
        \ENDFOR
    \ENDFOR
    \STATE \textbf{Return:} Preprocessed dataset $\hat{D}$.
\end{algorithmic}
\end{algorithm}

\subsection{Training}
\paragraph{Ablation study on training objective}
We evaluate the accuracy of the Entropy Tagger under two training objectives: classification and regression.
In the classification setting, the model is trained as a binary classifier to directly predict whether each token is low entropy. Accuracy is computed by comparing the predicted class label $\hat{y_i}\in\{0,1\}$, with the ground-truth label $y_i\in\{0,1\}$:
\begin{equation}
    \mathrm{Acc.}_{\mathrm{cls}} = \frac{1}{N}\sum_{i=1}^{N}\mathbf{1}[\hat{y_i} = y_i]
\end{equation}
In the regression setting, the model predicts a scalar entropy value $\hat{e_i}\in\mathbb{R}$. The ground-truth entropy value $e_i\in \mathbb{R}$ is also provided. We discretize both values into bins of width 0.3, capping the maximum bin value at 1.5, and evaluate accuracy by comparing the resulting discrete labels:
\begin{align}
    \mathrm{Bin}(x) &= \min\left( \left\lfloor \frac{x}{0.3} \right\rfloor \times 0.3,\ 1.5 \right) \\
    \mathrm{Acc.}_{\mathrm{reg}} &= \frac{1}{N} \sum_{i=1}^{N} \mathbf{1}\left[ \mathrm{Bin}(\hat{e}_i) = \mathrm{Bin}(e_i) \right]
\end{align}
Tab.~\ref{tab:accuracy_methods} demonstrates that the regression-based Entropy Tagger consistently underperforms the classification-based version in terms of accuracy across all three datasets. Consequently, we adopt the classification objective for training the Entropy Tagger.

\begin{table}[p]
\centering
\scalebox{0.85}{
\begin{tabular}{c|c|c|c}
\toprule
\multirow{2}{*}{Methods} & \textbf{MBPP} &\textbf{MBPP} & \textbf{HumanEval} \\ 
& \textbf{Validation} & \textbf{Test} & \textbf{Test} \\ \midrule
Regression  & 38.16  & 36.90     & 37.94          \\
Classification & 84.27  & 83.16     & 70.05     \\ \bottomrule
\end{tabular}
}
\caption{Accuracy of Entropy Tagger for different training objectives.}
\label{tab:accuracy_methods}
\end{table}

\paragraph{Details on Training Entropy Tagger}
During the training phase, we construct a binary classification MLP, and then, based on the threshold $\tau$, we map $y$ in $\hat{D}$ to True or False. If $y_i <\tau$, it is set to True, otherwise False. We then train using BCELoss and optimize with AdamW~\cite{loshchilov2017decoupled}. The hyperparameter settings are shown in the Tab.~\ref{tab:hyperparameter_settings}. Finally, the epoch with the highest accuracy on the MBPP validation split is selected as the Entropy Tagger.
\begin{table}[H]
\centering
\small
\begin{tabular}{l|l}
\toprule
\textbf{Hyperparameter} & \textbf{Setting} \\ \midrule
\# epochs & 100 \\
batch\_size    & 32      \\
lr  & 1e-4    \\
optimizer     & AdamW   \\
weight\_decay   & 2e-5  \\ \bottomrule
\end{tabular}
\caption{The hyperparameter settings for training the Entropy Tagger.}
\label{tab:hyperparameter_settings}
\end{table}

\subsection{Validation}
We use the MBPP test and HumanEval test as the test sets, representing the in-domain and out-of-domain scenarios, respectively. The test results are shown in Tab.~\ref{tab:accuracy_entropy_tagger}. 
The results show that the accuracy of the Entropy Tagger is consistent across different splits of the same dataset~(in-domain), achieving over 80\%. When applied across datasets~(out-of-domain), using the Entropy Tagger for prediction also achieves an accuracy of over 66.61\%, with an accuracy of 75.71\% at the $\tau=1.5$.

%% file: appendicies/implementation_details.tex
\section{Implementation details}
\label{app:implementaton_baselines}
\begin{figure*}[t]
    \centering
    \begin{subfigure}[t]{0.45\textwidth}
        \centering
        \includegraphics[width=\textwidth]{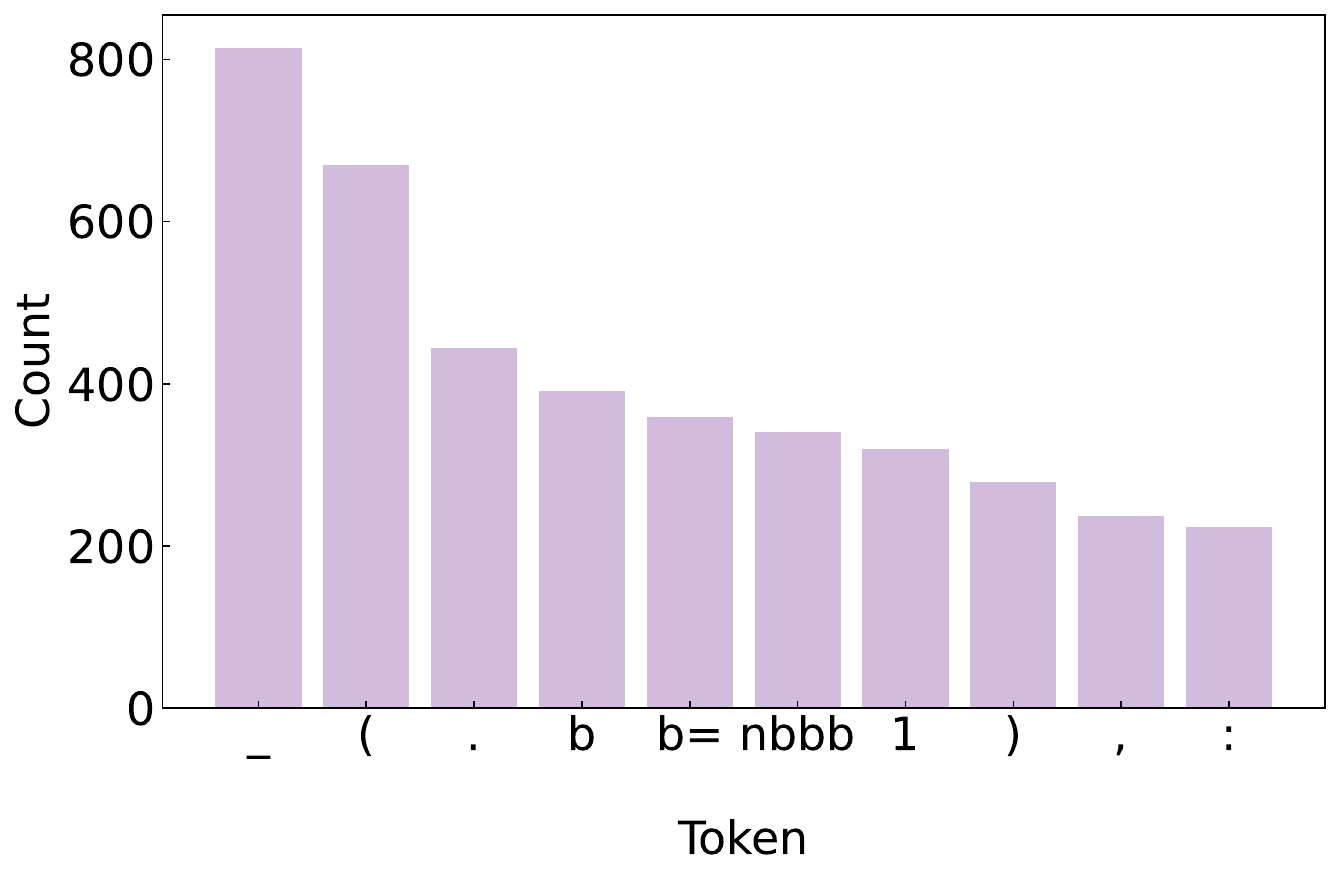}
        \caption{entropy threshold~$(\tau)$ = 0.3}
        \label{fig:image1}
    \end{subfigure}
    \hfill
    \begin{subfigure}[t]{0.45\textwidth}
        \centering
        \includegraphics[width=\textwidth]{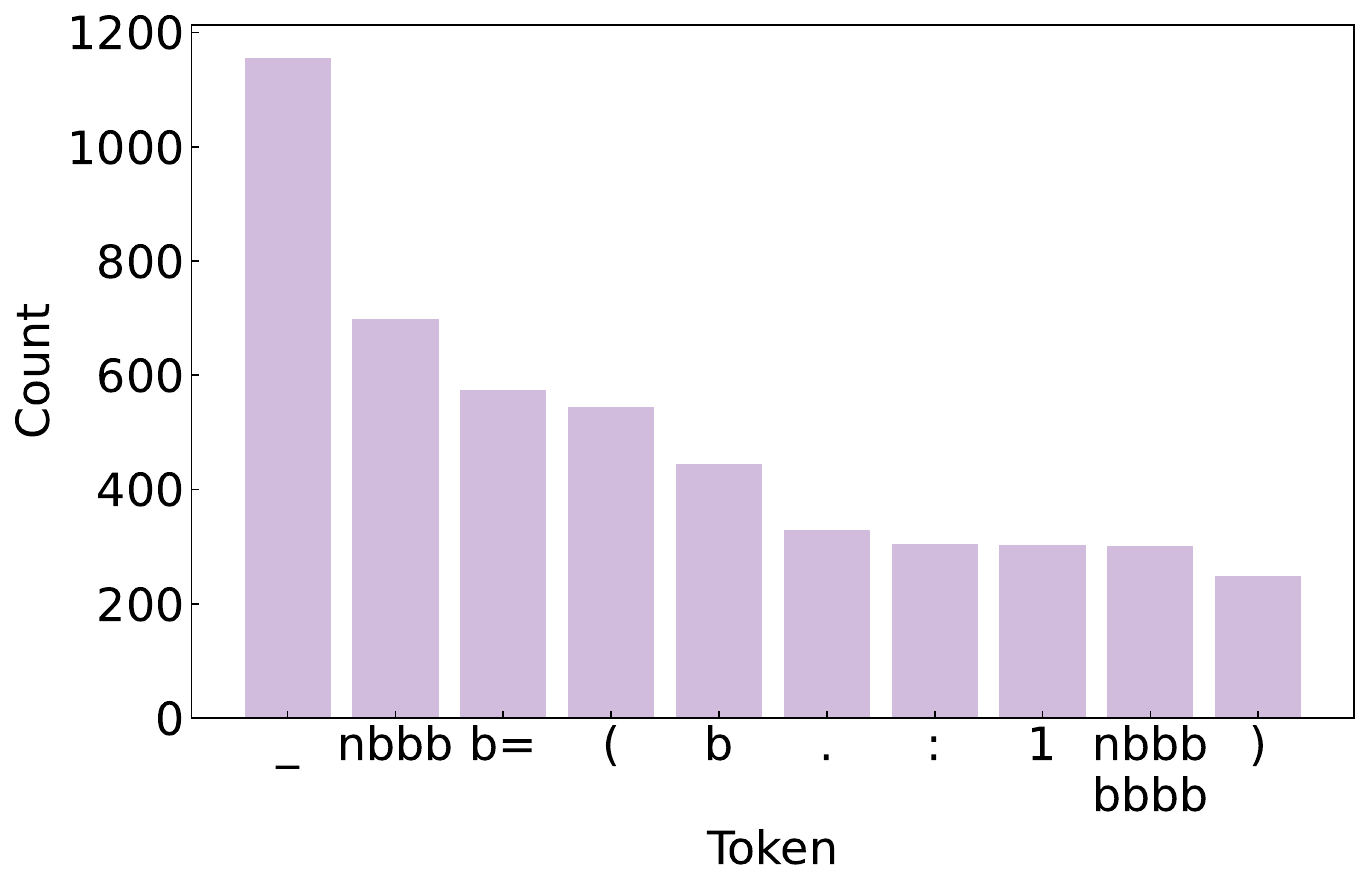}
        \caption{entropy threshold~$(\tau)$ = 0.6}
        \label{fig:image2}
    \end{subfigure}

    \begin{subfigure}[t]{0.45\textwidth}
        \centering
        \includegraphics[width=\textwidth]{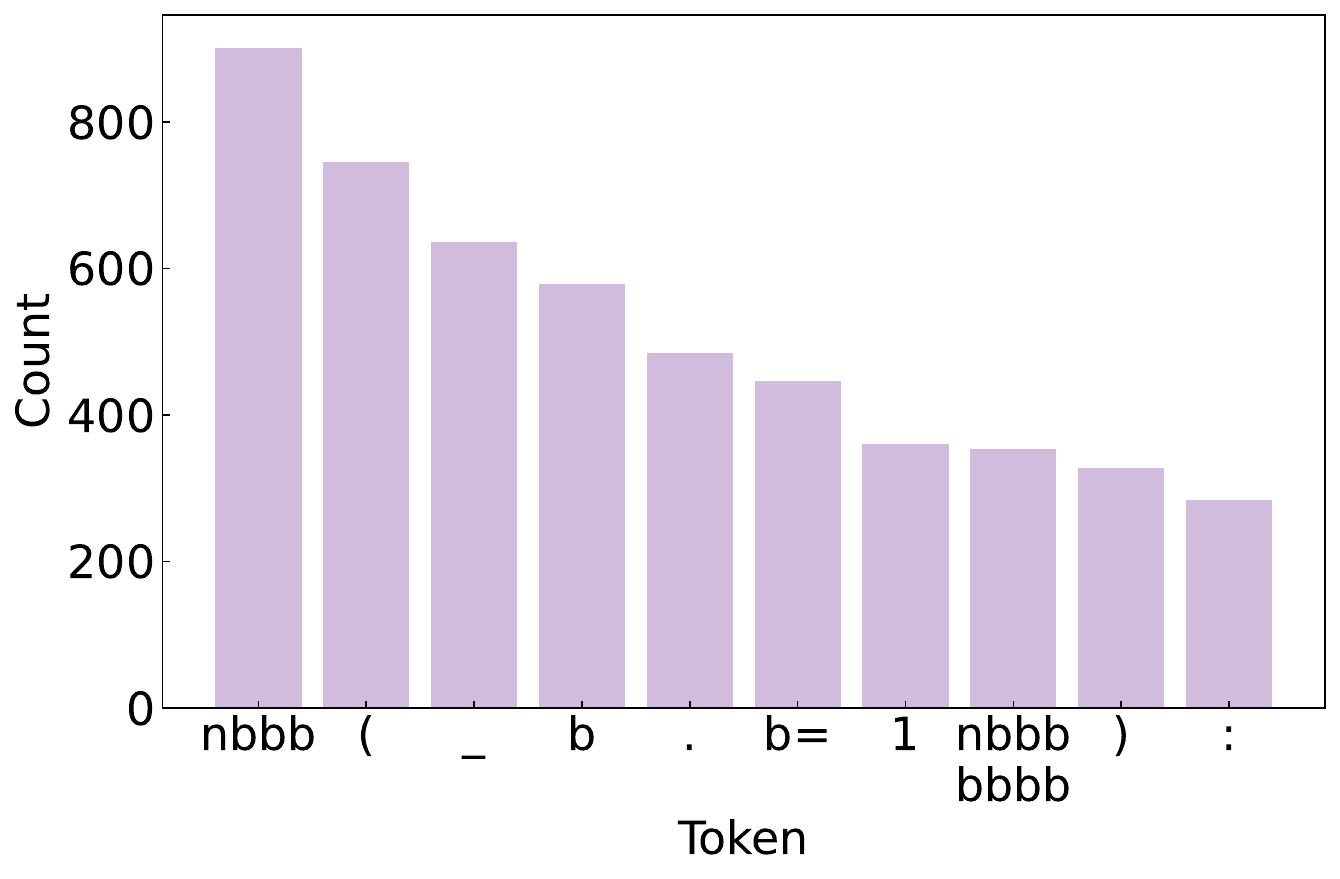}
        \caption{entropy threshold~$(\tau)$ = 0.9}
        \label{fig:image3}
    \end{subfigure}
    \hfill
    \begin{subfigure}[t]{0.45\textwidth}
        \centering
        \includegraphics[width=\textwidth]{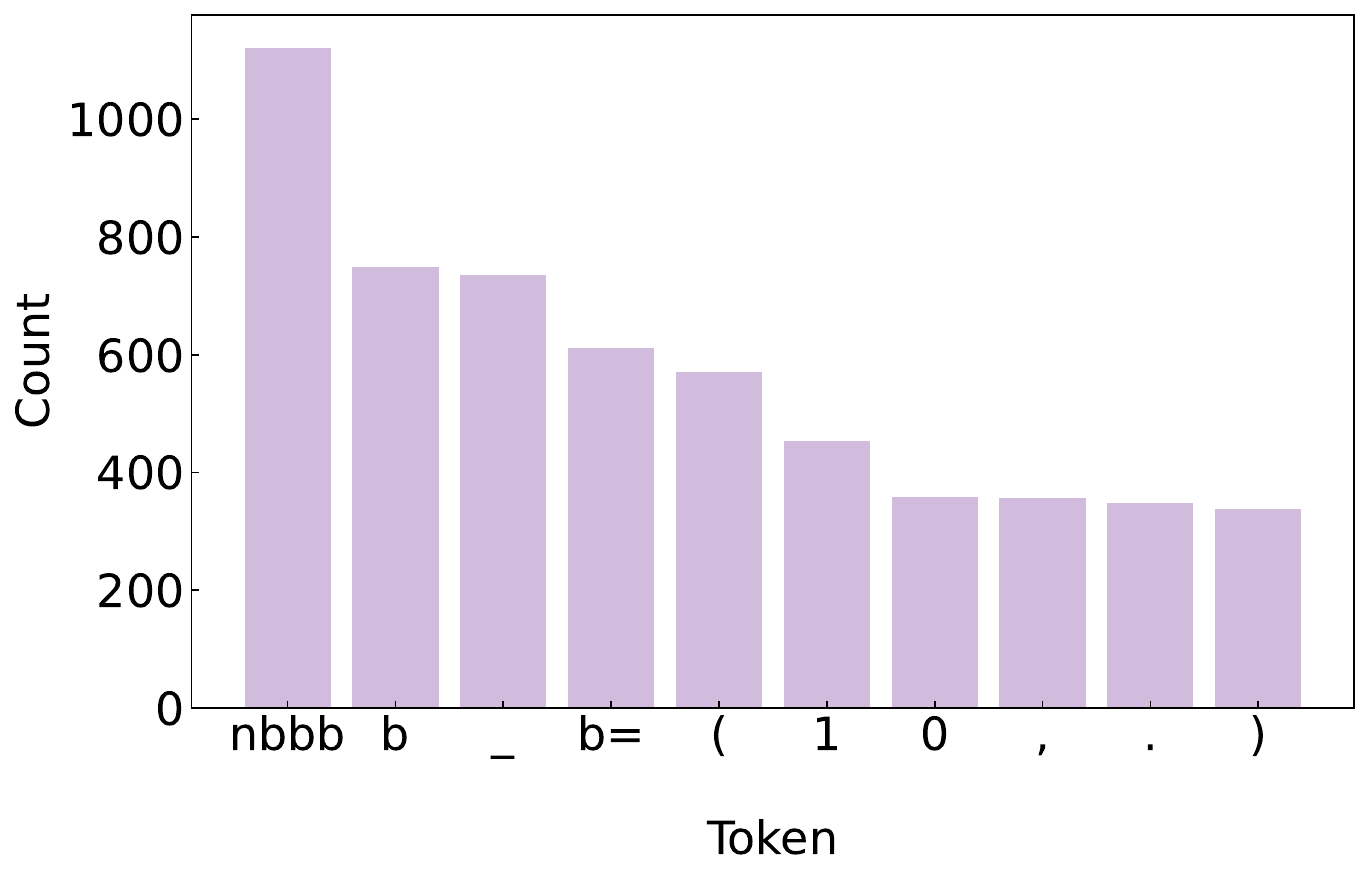}
        \caption{entropy threshold~$(\tau)$ = 1.2}
        \label{fig:image4}
    \end{subfigure}

    \caption{Top-K tokens most frequently classified as low entropy tokens.}
    \label{fig:top_k_tokens}
\end{figure*}
All methods can be implemented on a single \textit{NVIDIA A100-SXM4-40GB}. For Post-hoc methods and KGW, we follow the implementation provided in the ~\citealp{lee-etal-2024-wrote}. For EWD, we adopt the recommended hyperparameters from ~\citealp{lu-etal-2024-entropy}. However, to ensure a fair comparison, we use the same hash key in KGW for EWD. For SWEET, we use the settings recommended in the original paper. Since the Threshold Navigator automatically selects a fixed threshold, we report the averaged results across all thresholds for SWEET. As SWEET consider the trade-off between code generation ability and detectability, two results are reported for MBPP. We select the one with the highest AUROC. For IE, we report the result with the highest UES under the condition that Pass@1 is allowed to drop by up to 20\%. Detailed settings for each method on each dataset can be found in Tab.~\ref{tab:settings_implementation}.
\begin{table}
\centering
\scalebox{0.85}{
\begin{tabular}{llll}
\toprule
Dataset   & Method & $\gamma$ & $\delta$ \\ \midrule
\multirow{4}{*}{HumanEval} & KGW    & 0.25      & 3.0      \\
          & EWD    & 0.5      & 2.0      \\
          & SWEET  & 0.25     & 3.0     \\
          & IE     & 0.5      & 3.0      \\ \midrule
\multirow{4}{*}{MBPP}      & KGW    &  0.25     &  3.0     \\
          & EWD    &  0.5     &  2.0     \\
          & SWEET  &  0.5    &  2.0     \\
          & IE     &  0.25     &  3.0    \\ \bottomrule
\end{tabular}}
\caption{Detailed settings for each watermark methods.}
\label{tab:settings_implementation}
\end{table}

%% file: appendicies/detection_time.tex
\section{Computational Time Used Analysis}
To evaluate the computational efficiency of each method, we measure the total runtime required to complete evaluation on the \texttt{HumanEval} benchmark. Due to the variation in generated text lengths across different methods, all watermarking approaches are applied exclusively during the detection phase to ensure a fair comparison. Each method is evaluated three times under the same hardware conditions, and the average total runtime is reported. The results are summarized in Tab.~\ref{tab:time}.

\begin{table}[h]
\centering
\caption{Total runtime (in seconds) on the \texttt{HumanEval} benchmark for each method.}
\label{tab:time}
\begin{tabular}{ll}
\toprule
\textbf{Method} & \textbf{Total Time (s)} \\ \midrule
KGW & \phantom{0}55.86($\pm$ 10.67) \\
EWD & 118.83($\pm$ 11.82) \\
SWEET & 110.75($\pm$ 11.29) \\
IE &100.36($\pm$ 6.53) \\
\bottomrule
\end{tabular}
\end{table}

As shown in the Tab.~\ref{tab:time}, our method achieves the lowest total runtime, demonstrating its practical advantage in terms of computational efficiency.

%% file: appendicies/algoriths_for_threshold_navigator.tex
\section{Algorithms for Threshold Navigator}
The algorithmic details of Threshold Navigator are shown in Alg.~\ref{alg:threshold_navigator}. Given a prompt sequence, green list size, and search granularity, we begin by initializing the entropy threshold $\tau_0$ and computing the corresponding Watermark Ratio ($\mathrm{WR_0}$) and the number of green tokens $|S|_{G_{\tau_{0}}}$ under this threshold. (Lines 3-5). Then, we enumerate downward from the initial entropy threshold (e.g., 1.5) to lower values (e.g., 1.2, 0.9, 0.6, 0.3). For each new entropy threshold, we compute the updated Watermark Ratio~($\mathrm{WR}_{\tau_{i-1}})$ and green token count~($|S|_{G_{\tau_i}}$). (Lines 6-8) For every pair of adjacent entropy thresholds, we calculate the green token change ration $p$~(Line 9) and the Watermark Ratio change ratio $w$~(Line 10). The search stops when the condition $p>1$ and $w<1$ is met for the first time, and the previous entropy threshold is selected as the final threshold and returned~(Lines 11-13).
\begin{algorithm}[]
\small
    \caption{Threshold Navigator}
    \label{alg:threshold_navigator}
    \begin{algorithmic}[1]
        \STATE \textbf{Input:} \parbox[t]{\dimexpr\linewidth-\algorithmicindent}{
            prompt, $s_0, \ldots, s_{t-1}$ \\
            green list size, $\gamma \in(0,1)$ \\
            search granularity, $\Delta$ 
        }
        \STATE \textbf{Output:} $\hat{\tau}$ (final entropy threshold)
        \STATE Initialize $\tau_0$ (initial entropy threshold)
        \STATE $\hat{\tau} \gets \tau_0$
        \STATE Calculate $\text{WR}_0$ (Watermark Ratio) and $|S|_{G_{\tau_0}}$ (green token count).
        \FOR{$i = 1$ to $\lfloor \tau_0 / \Delta \rfloor$}
            \STATE $\tau_i \gets \tau_{i-1} - \Delta$
            \STATE Calculate $\text{WR}_{\tau_i}$ and $|S|_{G_{\tau_i}}$.
            \STATE $p \gets |S|_{G_{\tau_{i-1}}} / |S|_{G_{\tau_i}}$
            \STATE $w \gets \text{WR}_{\tau_{i-1}} / \text{WR}_{\tau_i}$
            \IF{$p > 1$ and $w < 1$}
                \STATE $\hat{\tau} \gets \tau_{i-1}$
                \STATE \textbf{break}
            \ENDIF
        \ENDFOR
        \STATE \textbf{return} $\hat{\tau}$
    \end{algorithmic}
\end{algorithm}

%% file: appendicies/analysis_of_low_entropy_token.tex
\section{Analysis on low entropy tokens}
\label{app:analysis_low_entropy}
We rank the frequency of tokens classified as low entropy token under $\gamma=0.25$ and $\delta=3.0$ across different entropy thresholds and report the top 10 tokens. To enhance clarity, we use "b" to represent spaces and "n" to represent newlines. The results are shown in Fig.~\ref{fig:top_k_tokens}. It can be observed that, despite varying entropy thresholds, certain tokens frequently appear as low entropy tokens, such as "\_", ".", ":", "1", "(", ")", spaces, newlines, and their combinations.